\documentclass[journal]{IEEEtran}

\usepackage[utf8]{inputenc} 
\usepackage[T1]{fontenc}    
\usepackage{url}            
\usepackage{booktabs}       
\usepackage{amsfonts}       
\usepackage{nicefrac}       
\usepackage{microtype}      
\usepackage{lipsum}

\usepackage{times}
\usepackage{epsfig}
\usepackage{graphicx}
\usepackage{amsmath}
\usepackage{amssymb}
\usepackage{multirow}

\usepackage{booktabs}
\usepackage{caption}
\usepackage[inline]{enumitem}

\usepackage[pagebackref=true,breaklinks=true,letterpaper=true,colorlinks,bookmarks=false]{hyperref}
\usepackage[nameinlink]{cleveref}

\begin{document}

\title{A Shared Representation for \\Photorealistic Driving Simulators}

\author{Saeed Saadatnejad,
        Siyuan Li,
        Taylor Mordan,
        Alexandre Alahi
\thanks{\copyright 2021 IEEE. Personal use of this material is permitted. Permission from IEEE must be obtained for all other uses, in any current or future media, including reprinting/republishing this material for advertising or promotional purposes, creating new collective works, for resale or redistribution to servers or lists, or reuse of any copyrighted component of this work in other works.}%
\thanks{Please cite as follows: S. Saadatnejad, S. Li, T. Mordan, and A. Alahi, "A Shared Representation for Photorealistic Driving Simulators," in IEEE Transactions on Intelligent Transportation Systems, DOI: 10.1109/TITS.2021.3131303.}%
\thanks{All researchers are with the VITA laboratory of EPFL, Switzerland. e-mail: (firstname.lastname@epfl.ch).}
}

\markboth{IEEE Transactions on Intelligent Transportation Systems}%
{IEEE Transactions on Intelligent Transportation Systems}

\maketitle

\begin{abstract}

A powerful simulator highly decreases the need for real-world tests when training and evaluating autonomous vehicles.
Data-driven simulators flourished with the recent advancement of conditional Generative Adversarial Networks (cGANs), providing high-fidelity images.
The main challenge is synthesizing photorealistic images while following given constraints.
In this work, we propose to improve the quality of generated images by rethinking the discriminator architecture. 
The focus is on the class of problems where images are generated given semantic inputs, such as scene segmentation maps or human body poses.
We build on successful cGAN models to propose a new semantically-aware discriminator that better guides the generator.
We aim to learn a shared latent representation that encodes enough information to jointly do semantic segmentation, content reconstruction, along with a coarse-to-fine grained adversarial reasoning.
The achieved improvements are generic and simple enough to be applied to any architecture of conditional image synthesis. 
We demonstrate the strength of our method on the scene, building, and human synthesis tasks across three different datasets. The code is available \href{https://github.com/vita-epfl/SemDisc}{https://github.com/vita-epfl/SemDisc}.
\end{abstract}


\IEEEpeerreviewmaketitle

\section{Introduction}
\label{sec:intro}


\IEEEPARstart{S}{afety} is the primary concern when developing autonomous vehicles (AVs).
For example, a wrong action in an unexpected situation can lead to a collision with a pedestrian, which is not negligible \cite{bouhsain2020pedestrian, parsaeifard2021decoupled}.
Yet, strictly evaluating AVs in the real world is not a realistic nor a safe option. 
Some argue that an AV should be tested millions of miles in challenging situations to demonstrate its performance \cite{karla2016drivingsafety}.
Besides its extensive required time and costs, it is impossible to cover all rare cases.
Simulations can play a significant role in overcoming these issues \cite{payalan2020surveysim}.
By synthesizing images, we are able to not only evaluate the performance of AVs but also improve the performance of current deep networks leveraging the abundant amount of data \cite{vondrick2016generating, gaidon2016virtualworlds,zheng2019jointdg,liu2018posetransferrable,beery2020rareClass,saadatnejad2021sattack}.

Researchers have investigated two paradigms: model-based and data-driven simulators. The former is based on physics laws and computer graphics, such as Carla \cite{dosovitskiy2017carla}.
It needs high-fidelity environmental models to create indistinguishable images, which is highly expensive.
The latter learns to effectively generate the images from examples \cite{wang2018pix2pixhd,amini2020mitsimulator,yang2020surfelgan}. In this work, we tackle the semantically-driven image synthesis task: given a semantic mask (\textit{e.g.}, human body poses, or scene segmentation masks), we aim to generate a realistic image with the same semantics.

\begin{figure*}[!t]
    \begin{center}
        \includegraphics[width=0.85\textwidth]{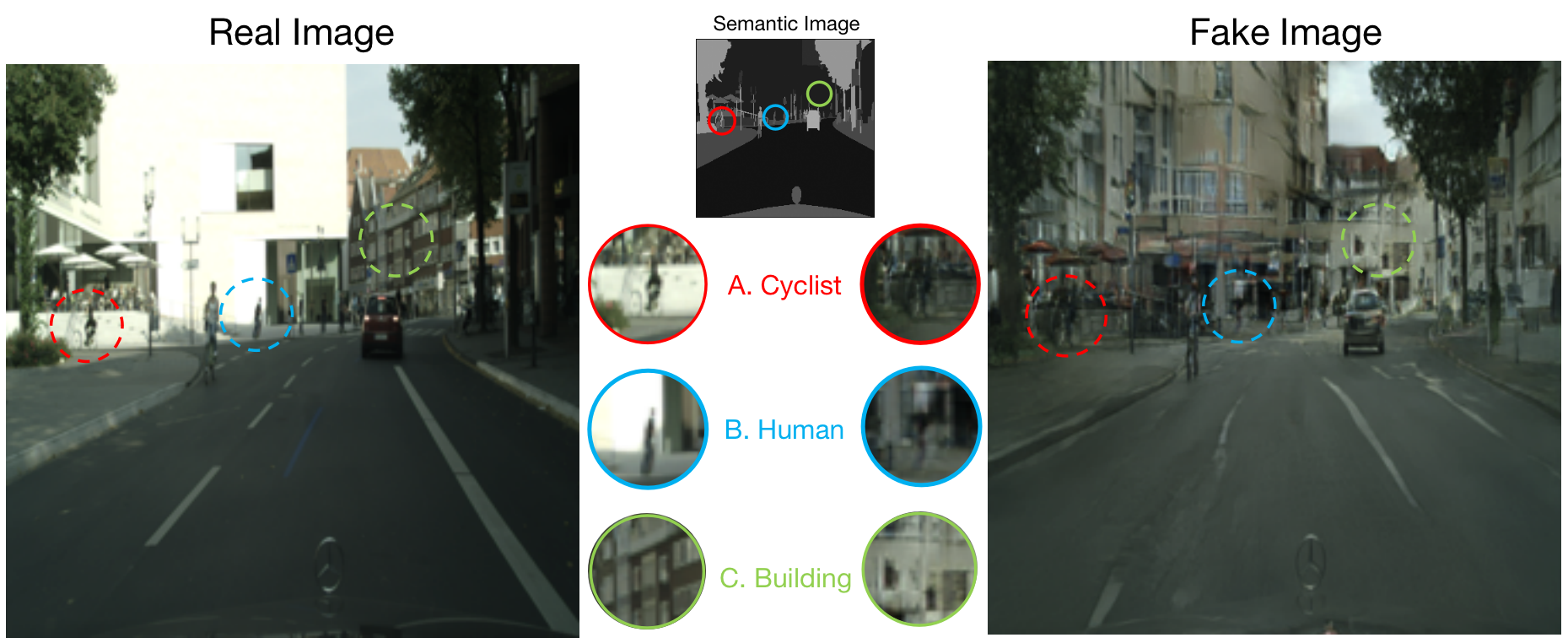}
    \end{center}
    \caption{Given the appropriate semantic map, the network is supposed to synthesize a realistic image with the desired semantic. Although a fake image may look realistic from a global view, two problems remain: some semantics are not followed (A and B) and fine-grained details reveal the fake one (C).}
    \label{fig:pull}
\end{figure*}

Photorealistic image synthesis is a notoriously difficult task due to a high dimensional output space and an ill-posed objective. 
It is commonly done with conditional Generative Adversarial Networks (cGANs) \cite{isola2017pix2pix, wang2018pix2pixhd, park2019gaugan, siarohin2018deformable}. 
However, state-of-the-art approaches cannot always provide enough supervision to the generator.
As a solution, some provide a structured semantic description as another input to the discriminator.
The discriminator of the cGAN is in charge of classifying the whole image as real or synthetic conditioned on the specified semantic input, hoping to learn the joint distribution of (image, segmentation).
Yet, learning to make generated images realistic leads to not perfectly following the semantic content, especially for some small or rare objects, as shown in \Cref{fig:pull}.
Another solution to attain high fidelity images is adding conditional matching losses in the pixel space. This is too strict as only the high-level description needs to be followed. Indeed, the discriminator is bypassed, and the generator is directly supervised by the content reconstruction.
Finally, there also exists another problem in the main adversarial task. The discriminator gives the same weight to all image regions and does not learn a specialized network for the texture of a specific semantic class.
For instance, what makes a car real might differ from what makes a road real.

We argue that the task of the discriminator, classifying a real/fake image, is closely related to having the capacity to understand its content \textit{e.g.}, recognizing semantic, and compressing it. Hence, we ask the discriminator to perform three tasks: (1) image segmentation, to verify the loyalty of the generated image and the requested label, (2) reconstruction task, aiming at the conceptual understanding of the semantics, and (3) coarse-to-fine grained adversarial task trying to distinguish between fake and real in a class-specific manner.
Since all these tasks share some useful information in the pixel-domain, we propose to learn them within the same representation as the adversarial supervision.
In this paper, we learn a shared latent representation that encodes enough information to jointly do semantic segmentation, content reconstruction, along with a coarse-to-fine-grained assessment. This leads to a more semantic-consistent output, more stable training, and more details in the images. 

\begin{figure*}[!t]
    \begin{center}
    \includegraphics[width=0.8\textwidth]{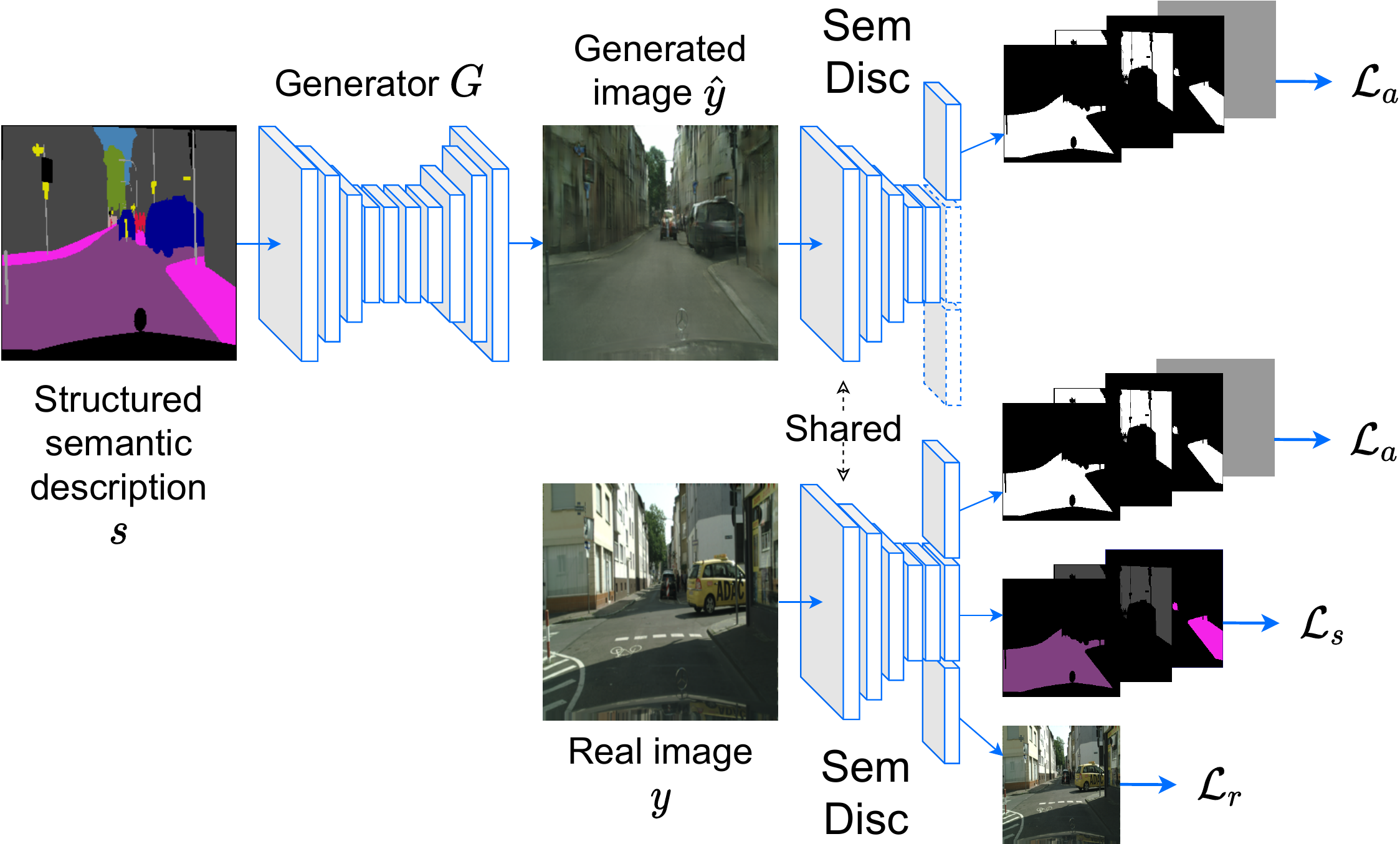}
    \end{center}
       \caption{Conditional GAN training with semantic guiding. SemDisc has three heads. The first head provides several maps gated by semantic masks corresponding to the structured high-level description of the target output to focus learning on relevant areas. In the second head, the semantic is also leveraged to compute a semantic loss, matching the given constraint in a suitable space rather than a pixel one. The reconstruction head is supposed to reconstruct the image back, matching the texture in the pixel domain. The first head is trained on real and generated images, but other heads are trained only on real images.}
    \label{fig:method}
\end{figure*}

Our main contribution is learning a shared representation to provide correct supervisions for the generator. This is performed by a new architecture for the discriminator called SemDisc, in a multi-task learning approach, which is shown in Figure \ref{fig:method}. The discriminator consists of three heads: the first head (semantics) forces the generator to follow the semantics explicitly, the second head (reconstruction) reconstructs the image back, acting as a regularizer to the training process, and the third head (coarse-to-fine adversarial) modifies the loss function to maintain coarse-to-fine grained details in the generated image. Finally, we introduce a trick to stabilize the training process.

The improvements we present in this paper are generic and simple enough that any architecture of cGAN could benefit from a conversion from a regular discriminator to a structured semantic one.
Interestingly, as only the discriminator is modified, it should be independent of the particular generator architecture used and should also be complementary to any approach based on generator enhancement, \textit{e.g.}, \cite{siarohin2018deformable}, \cite{wang2018pix2pixhd}, \cite{park2019gaugan}.

Since the discriminator is used during training only, it is noticeable that all the changes we apply do not bring any run-time overhead, both in forward time and memory footprint.
All effects happen through better learning of the model, thanks to the shared representation learning, compared to modifying the generator network, \textit{e.g.}, by adding additional capacity to it that might impact the forward pass.

Finally, to show how our approach can influence the training of AVs, we share an in-depth analysis of our model on three image generation datasets that are related to transportation. This covers car-view image synthesis and building image synthesis from segmentation maps and human image synthesis from body poses (keypoints).

\section{Related Work}
\label{sec:rel}

\paragraph{\textbf{Image generation}}
Most recent deep learning methods for image synthesis use Generative Adversarial Networks (GANs) \cite{goodfellow2014generative} and Variational AutoEncoders (VAEs) \cite{kingma2013vae}.
GANs use two separate networks, a generator, and a discriminator that jointly optimize exclusive objectives.
In this process, the generator learns to generate more realistic images, and the discriminator learns to distinguish between real and fake images more accurately.
VAEs are another type of generative models that rely on probabilistic graphical models.
Although they have been shown to disentangle features, the generated images are usually not as realistic as those from GANs \cite{esser2018vunet}.
In this paper, we mainly consider GANs.

Several methods have modified the design of the generator of GANs to get better results.
Using mapping networks, adaptive instance normalization (AdaIN) \cite{karras2018stylegan} and spatially adaptive normalization (SPADE) \cite{park2019gaugan} are among successful ideas in improving its architecture.
These kinds of improvements have recently led to stunning results in the generation of natural images \cite{brock2018biggan} or human faces \cite{karras2017progressive,karras2018stylegan}.
%
Moreover, it has been shown that these realistic generated images could be used as data augmentation in other tasks to improve accuracy, \textit{e.g.}, in person re-identification \cite{liu2018posetransferrable}, \cite{zheng2019jointdg}, semantic segmentation \cite{zheng2020forkgan}, \cite{romera2019bridging} and even inspection of defect railway fasteners \cite{liu2021railway}.

\paragraph{\textbf{Conditional image generation}}
Conditional GANs (cGANs) generate images from other images or high-level descriptions of the desired outputs.
Applications can be various, as exemplified by pix2pix \cite{isola2017pix2pix}, which applies the image-to-image translation approach to a wide range of computer vision problems.
More recently, realistic results have been obtained in the generation of city scenes using semantic maps \cite{wang2018pix2pixhd}, \cite{park2019gaugan}, \cite{shaham2021asapnets}, and even talking head videos from few examples \cite{zakharov2019fewshottalkinghead}.

To improve the quality of generated images, some added an auxiliary classification task \cite{acgan}, 
some tried to find and modify the regions of interest by means of attention maps \cite{tang2019multichannelattention}, \cite{alami2018unsupervisedattentionguided} 
and recently, some used pre-trained segmentation networks \cite{cherian2019sem}.
However, in \cite{chen2017crn}, they showed that merely adding the segmentation loss (pixel-wise cross-entropy loss) leads to unstable training with many artifacts.
Indeed, they defined a baseline with an additional term in the loss function that when the synthesized image is given as input to a pretrained semantic segmentation network, it should produce a label map close to the input semantics.

\paragraph{\textbf{Conditional human image generation}}
In spite of realistic results in face image generation, human image synthesis is far from looking real since images need fine details of all body parts for a synthesized image to be considered as real.
The problem becomes harder in conditional human image synthesis, where the model has to preserve the identity and texture of the conditioned image.
One major issue is large body deformations caused by people's movements or changes in camera viewpoint.
Several ideas have been developed. \cite{pumarola2018unsupervised} added a pose discriminator, \cite{siarohin2018deformable} introduced deformable skip connections in its generator and used a nearest neighbour loss, \cite{ma2018bodyroi7}, \cite{ma2017pg2} disentangled foreground people from background to transform them into the new pose while trying to have a background close to the source image.
\cite{saadatnejad2019pedestrian} learned a latent canonical view of a pedestrian in order to generate in any pose.
\cite{dong2018softgated} designed a soft-gated Warping GAN to address the problem of large geometric transformations in human image synthesis.
\cite{chan2018everybodydance}, \cite{wang2018vid2vid} trained a personalized model for each person, and \cite{wang2018fewshotvid2vid} leveraged a few-shot learning approach needing few images from a new person to refine the network at test time.

\paragraph{\textbf{Discriminator in image generation}}
The architecture of the discriminator plays a role in the quality of generated images through the learning of the generator.
Patch-wise discriminators (PatchGANs) have outperformed global ones with full-image receptive fields for both neural style transfer \cite{li2016patchgan} and conditional image generation \cite{isola2017pix2pix}.
Although the discriminator is often discarded after training, some methods leverage the information it learns.
\cite{wang2018pix2pixhd} yields high-quality images by having multiple discriminators at different resolutions, and \cite{chan2018everybodydance} uses two separate networks for synthesizing full-body and face.
\cite{liu2019collaborative} improves the quality of generated images and prevents mode collapse by leveraging the information stored in the discriminator and reshaping the loss function of GAN during image synthesis.

\cite{li2018semgradgan} also uses semantics to guide the discriminator but in a setup of image translation between domains (real $\Leftrightarrow$ virtual).
Thus, they need images as input while we need semantics.
Their approach is restricted to cases when the segmentation label maps cover the whole image.
However, ours, by modifying the masking process and adding the coarse layer (responsible for making the whole image realistic), can work in all non-complete semantic maps such as human image synthesis.
Moreover, we provide a multi-task learning approach with an added semantic matching head.


Recently, some others tried to leverage a U-net architecture in their discriminator.
\cite{schonfeld2020unetD} provided detailed per-pixel feedback to the generator while maintaining the global context in an unconditional setting (without semantics). 
\cite{liu2019featurePyramidD} modified the discriminator and defined a semantic alignment score map derived by multiplying activations of different layers of the discriminator with the ground truth label map. This strict constraint, which acts as a regularizer, could slightly improve the scores.

\section{Method}
\label{sec:method}

We propose a multi-task learning approach to address conditional Generative Adversarial Network (cGAN) training for general-purpose image synthesis. We include structured semantic information to guide learning in order to focus more on meaningful regions of images.
We build on successful cGAN models \cite{park2019gaugan}, \cite{siarohin2018deformable} and propose to add the following appropriate supervisions: (i) biasing the discriminator toward semantic features, (ii) training a semantic matching, and (iii) adding a novel reconstruction loss which will subsequently influence the learning of the image generator network.

\subsection{Overview of the approach}
\label{sec:overview}

Our model is composed of a main network $G$ generating an image $\hat{y} = G(s)$ from a structured semantic description $s = (s_1, \ldots, s_K)$ over $K$ feature maps (\textit{e.g.}, class masks or heatmaps of keypoints) of the desired output, as depicted in \Cref{fig:method}.
During learning, examples consist of pairs $(y, s)$ of real images $y$ and their corresponding semantic descriptions $s$.
After training, the distribution of generated images $\hat{y} = G(s)$ is expected to be similar to the distribution of $y$ as they should share the same underlying semantic structures $s$.
However, it is not easy to handcraft a loss function to assess the quality of the outputs $\hat{y}$ of $G$.
For this, a discriminator network $D$ is concurrently trained with it to act as a proxy loss, both networks competing to optimize exclusive loss functions in an adversarial minimax game \cite{goodfellow2014generative}.

As illustrated in \Cref{fig:method},
our approach is different from common conditional GAN discriminators in which the image and the semantic map are concatenated. Indeed, the discriminator $D$ takes as input an image $x$, and its semantic description $s$ is applied in the loss function.
The image $x$ can either be generated by $G$ (in which case $x=\hat{y}$) or be a real image ($x=y$), and $D$ is trained to identify this, through minimization of an adversarial loss $\mathcal{L}_a$, a semantic matching loss $\mathcal{L}_s$ and a new reconstruction loss $\mathcal{L}_r$.
At the same time, the generator $G$ learns to generate images that both are realistic and match the input constraints.

In order to generate realistic images, the generator $G$ learns to fool the discriminator $D$ by maximizing its loss $\mathcal{L}_a$.
Usually, the training of cGANs does not leverage all the semantic content of the description $s$.
We suggest that properly incorporating structured semantics into the training should result in generated images with better details around these semantic features.
For this, we modify the discriminator network $D$ and its associated loss function $\mathcal{L}_a$, which will impact the training of the generator $G$, as detailed in \Cref{sec:adversarial}.

The second objective to be optimized by the generator is having images matching their semantic descriptions.
It is usually achieved by training $G$ with the guidance of $D$.
However, it only uses the description $s$ as input to $D$ to check whether the image matches it.
To solve this issue, we split the task of $D$. The new head explicitly minimizes the semantic loss $\mathcal{L}_{s}$, described in \Cref{sec:segmentation}.
Moreover, to regularize the training, we define a novel reconstruction loss $\mathcal{L}_{r}$, described in \Cref{sec:reconstruction}.

The complete loss function $\mathcal{L}_G$ to be minimized by the generator network $G$ is therefore
\begin{equation}
    \mathcal{L}_G = -\mathcal{L}_a + \lambda_s\mathcal{L}_{s} + \lambda_r\mathcal{L}_{r},
    \label{eq:g_loss}
\end{equation}
where $\lambda_s$ and $\lambda_r$ are weighting coefficients between those loss terms. For the discriminator, it is similar with increasing the adversarial loss
\begin{equation}
    \mathcal{L}_D = \mathcal{L}_a + \lambda_s\mathcal{L}_{s} + \lambda_r\mathcal{L}_{r}.
    \label{eq:d_loss}
\end{equation}
Note that in our approach, $D$ is composed of three heads $D_a, D_s, D_r$ which share all layers except the last convolution layer, which will be described later.

\subsection{Coarse-to-fine adversarial head}
\label{sec:adversarial}

Our discriminator architecture is based on PatchGAN's one \cite{li2016patchgan}, whose output consists of a feature map where the score at each location indicates whether the corresponding input image patch is real or generated.
PatchGAN discriminator $D_{patch}$ is trained with a classification cross-entropy loss function $\mathcal{L}_{D_{patch}}$. We removed the semantics from its input which leads to
\begin{equation}
\begin{split}
        \mathcal{L}_{D_{patch}}(\hat{y}, y) = & 
        \mathbb{E}_{y} \left[ -\log\left( D_{patch}(y) \right) \right] \\ & + \mathbb{E}_{\hat{y}} \left[ -\log\left( 1 - D_{patch}(\hat{y}) \right) \right].
    \label{eq:patchgan_loss}
\end{split}
\end{equation}
However, instead of having a single output map globally classifying images, we here use multiple ones and force each of them to focus on a different semantic feature described by $s$.

Specifically, as illustrated in \Cref{fig:method}, for a structure $s$ with $K$ channels, the coarse-to-fine adversarial head of the discriminator $D_a$ outputs $K+1$ maps $D_a(x) = \left(D_{a_0}, D_{a_1}, \ldots, D_{a_K}\right)$.
The first map, $D_{a_0}$, handles the whole foreground objects described by the full tensor $s$ at a coarse, global scale.
Then, each map $D_{a_k}$ of the remaining $K$ ones corresponds to a given localized semantic feature $s_k$, in order to model fine-grained details associated with this feature.

To guide the learning of the various fine-grained prediction heads toward their corresponding semantic regions, semantic masks $M_k(s_k)$ are designed from the features $s_k$ to indicate their locations within images.
Note that the exact way semantic masks $M_k$ are obtained from the features $s_k$ depends on the type of their structures and is described in \Cref{sec:exp} for each dataset separately.    
The classification loss $\mathcal{L}_{D, k}$ used to train the branch $k$ is then element-wise multiplied with its associated mask $M_k$ to select spatial areas that are relevant for the semantic feature attended to.
Thus, backpropagation happens on the selected elements and their surroundings only, so that other regions of images not related to this feature do not affect the training.
By explicitly attending to different semantic areas, it should be easier for the discriminator to focus on local details not easily captured by a global view on the image so that the generator learns to refine them.
Regarding the coarse scale, the mask $M_0(s)$ is defined as covering the whole image and used in the same way.
The complete loss function $\mathcal{L}_a$ for this head of the discriminator $D_a$ is then the weighted sum, with equal weight given to the coarse loss than to all other fine-grained ones as these should only refine the first one,
\begin{equation}
    \mathcal{L}_a = \mathcal{L}_{a, 0} + \sum_{k=1}^{K} \frac{1}{K} \mathcal{L}_{a, k},
\end{equation}
where each term $\mathcal{L}_{a, k}$ is defined by the masked\footnote{extending the notation with $s_0=s$.} version of the PatchGAN loss function from \Cref{eq:patchgan_loss}:
\begin{equation}
    \begin{split}
        \mathcal{L}_{a, k} &= \mathbb{E}_{y,s} \left[ -\log\left( D_{a,k}(y) \right) \odot M_k(s_k) \right] \\
            &\quad+ \mathbb{E}_{\hat{y},s} \left[ -\log\left( 1 - D_{a,k}(\hat{y}) \right) \odot M_k(s_k) \right],
    \end{split}
\end{equation}
and is normalized by the number of pixels contained in the mask $M_k(s_k)$.
Note that when learning the generator $G$ by maximizing $\mathcal{L}_a$ (\Cref{eq:g_loss}), only the expectation over $\hat{y}$ is relevant, the other term being independent of $G$.

\subsection{Semantic matching head}
\label{sec:segmentation}

We argue that a perceptual loss commonly used to match the generated images with the target ones (\textit{e.g.}, \cite{park2019gaugan}) impose too strict requirements because an optimization in the pixel space would guide the model to yield these specific target images, while they should represent possible desired outputs only.
Therefore, we introduce a semantic matching loss function to relax these constraints and instead match images in a semantic space which yields more diversity and less blurring in synthesized images.

For this, we add another head to the discriminator that predicts the semantic description $s$ of the input image ($y$ or $\hat{y}$).
Specifically, as illustrated in \Cref{fig:method}, for a structure $s$ with $K$ channels, the semantic matching head of the discriminator $D_s$ outputs $K$ maps each matching the correspondent constraint.
The semantic loss is defined as a cross-entropy function between the upsampled outputs of this head ($D_s$) and the real semantic maps. This upsampling is necessary to match the size of the input semantic maps.

\subsection{Reconstruction head}
\label{sec:reconstruction}

To regularize the training, some use a regression loss, \textit{e.g.}, a $L_1$ loss \cite{isola2017pix2pix}. 
However, it suffers from blurriness and lowering the diversity of generated images.

We introduce a novel reconstruction loss $\mathcal{L}_{r}$, as another head of the discriminator.
This head acts as a regularizer: $\mathcal{L}_{r} = |f_{up}(D_r(y)) - y|$ where $f_{up}$ stands for the upsampling function to match the image size.
Note that this head is trained only on real images.

\subsection{Stabilizing the training}
\label{sec:stablizing}

Employing the defined loss function and following the routine training process (training $D$ with real and fake images and training $G$ with fake images) leads to unstable training, which will be discussed in this section. 

Take the joint distribution of training data as $p^*(x,s)$, the goal is to find an approximate joint distribution $p_{\theta}(x,s)$. The full objective function was defined in \Cref{eq:g_loss} and \Cref{eq:d_loss}. For simplicity, we ignore the reconstruction loss here and therefore the objective function is as follows:

\begin{equation}
    \underbrace{d(p^*(x), p_{\theta}(x))}_{\textcircled{a}} - \underbrace{{\mathbb{E}}_{p^*(x,s)}[\log(q(s|x)]}_{\textcircled{b}} - \underbrace{\mathbb{E}_{p_{\theta}(x,s)}[\log(q(s|x)]}_{\textcircled{c}} 
\end{equation}

where $d(p^*(x), p_{\theta}(x))$ is the Jensen-shanon divergence and $q(s|x)$ is the semantic matching head.
The term \textcircled{a} corresponds to common adversarial training loss that $G$ tries to minimize it and $D$ maximize.
The terms \textcircled{b} and \textcircled{c} which correspond to real and fake images respectively, can be written as follows:
\begin{equation}
    \textcircled{b} = \mathbb{E}_{p^{*}(x)}[H_{p^{*}(x)}(s|x)] + \mathbb{E}_{p^{*}(x)}[KL(p^{*}(s|x)||q(s|x))]
    \label{eq:real_auxilliary}
\end{equation}
\begin{equation}
    \textcircled{c} = \mathbb{E}_{p_{\theta}(x)}[H_{p_{\theta}(x)}(s|x)] + \mathbb{E}_{p_{\theta}(x)}[KL(p_{\theta}(s|x)||q(s|x))]
    \label{eq:fake_auxiliary}
\end{equation}

To derive \Cref{eq:real_auxilliary}, consider the following:
\begin{equation} \label{eq1}
\begin{split}
    & - \mathbb{E}_{p^{*}(x)}[H_{p^{*}(x)}(s|x)] = \mathbb{E}_{p^{*}(x)}[\log p^{*}(s|x)] \\
            &= \mathbb{E}_{p^{*}(x,s)}[\log\frac{p^{*}(s|x)q(s|x)}{q(s|x)}] \\
            &= \mathbb{E}_{p^{*}(x,s)}[\log q(s|x)] + \mathbb{E}_{p^{*}(x)}\mathbb{E}_{p^{*}(s|x)}[\log\frac{p^{*}(s|x)}{q(s|x)}] \\
            &= \mathbb{E}_{p^{*}(x,s)}[\log q(s|x)] + \mathbb{E}_{p^{*}(x)}[KL(p^{*}(s|x)||q(s|x))].
\end{split}
\end{equation}
If we replace $p^{*}$ with $p_\theta$, \Cref{eq:fake_auxiliary} is derived.

For real images, \Cref{eq:real_auxilliary} is optimized. Its first term is zero and the second term makes $q(s|x)$ a good approximation of the real distribution $p^*(s|x)$.

For fake images, \Cref{eq:fake_auxiliary} is optimized by two steps: (i) training $G$ while $D_s$ is frozen which pushes $p_{\theta}(s|x)$ towards $q(s|x)$,
(ii) training $D_s$ while keeping $G$ frozen which pushes $q(s|x)$ towards $p_{\theta}(s|x)$.
The effect of the second step is counter-intuitive. $q(s|x)$ is used as an intermediary distribution to help $p_{\theta}(s|x)$ be a variational approximation of $p^*(s|x)$ while this pushes $q(s|x)$ towards $p_{\theta}(s|x)$. 
In order to avoid that issue, the semantic matching head ($D_s$) is not trained on fake images leading to more stable training.

Another point in the training is the initialization of the semantic matching head in order to avoid the misguidance of the generator. In other words, in the beginning, $G$ is not receiving any gradients from the semantic matching head since it is not trained well.

\section{Experiments}
\label{sec:exp}

We evaluate our model on two sets of experiments in different domains, showing the benefits of our proposed method on image synthesis from semantics: scene synthesis from segmentation maps (both car-view scenes and city buildings) and human synthesis from keypoints.

\subsection{Scene synthesis from segmentation maps}
\label{sec:scenes}
 
\paragraph{Datasets}
For the task of generating scene images from segmentation maps, we use two different datasets.
The CMP Facades dataset \cite{tylecek2013facades} has $606$ images of different resolutions of buildings with their $c=12$ class semantic masks.
We use the same split as \cite{park2019gaugan}, composed of $400$ training, $100$ validation and $106$ test examples.
Cityscapes \cite{cordts2016cityscapes} is a dataset of road scenes, with $2,975$ images in training and $500$ in validation.
Semantic annotations are segmentation maps in $c=19$ classes.
All images of those datasets are resized to a resolution of $256 \times 256$.

\paragraph{Implementation details}
We use the same generator as \cite{park2019gaugan} and modify their two discriminators (in two different resolutions), \emph{i.e.}, the last convolution layer (contains $512$ kernels of size $4 \times 4 \times 1$) is replaced by three convolution layers, and each outputs multiple feature maps instead of a single feature map (contains $512$ kernels of size $4 \times 4 \times ((c+1)+c'+3)$). The $c+1$ feature maps are related to the coarse-to-fine adversarial head, and $c'$ and $3$ are for semantic matching and the reconstruction heads, respectively.
Similar to them, we replaced the LS-GAN loss in \Cref{eq:patchgan_loss} with the hinge loss.
$\lambda_s$ and $\lambda_r$ are assigned to $1.0$.
Learning is conducted with Adam optimizer with a learning rate of $0.0002$, momentum parameters $\beta_1 = 0.5$, $\beta_2 = 0.999$ for $200$ epochs.
In the first $100$ epochs, the generator is not trained with semantic matching and reconstruction heads, but for the next $100$ epochs, it is trained by all heads.
Moreover, we linearly decay the learning rate to $0$ from epoch $100$ to $200$.
Finally, the batch size is 32, and the hardware that we used contains $4$ 32GB V100 NVIDIA GPUs.

Semantic masks $M_k(s_k)$ are simply the downsampled segmentation masks defined by $s_k$.
We keep the perceptual loss and feature matching loss, which were used in the baseline.
Results of SPADE baseline \cite{park2019gaugan} are obtained by re-training their model with their hyper-parameters publicly available.\footnote{\url{https://nvlabs.github.io/SPADE/}}

\begin{figure*}[!t]
    \begin{center}
        \makebox[0.18\textwidth][c]{(a) Ground truth}
		 \makebox[0.18\textwidth][c]{(b) Semantic input}
		 \makebox[0.18\textwidth][c]{(c) SPADE \cite{park2019gaugan}}
		 \makebox[0.18\textwidth][c]{(d) Ours} \\
        \smallskip
        \includegraphics[width=0.18\textwidth]{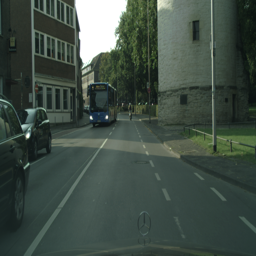}
         \includegraphics[width=0.18\textwidth]{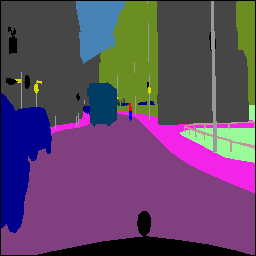}
         \includegraphics[width=0.18\textwidth]{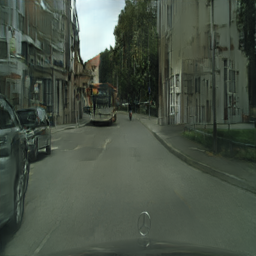}
         \includegraphics[width=0.18\textwidth]{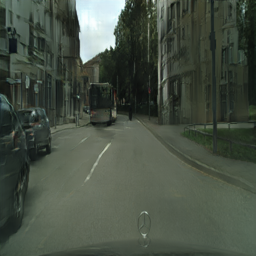} \\
        \smallskip
        \includegraphics[width=0.18\textwidth]{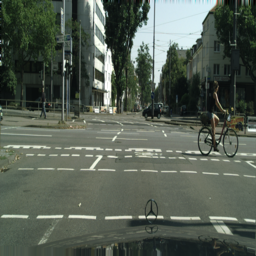}
         \includegraphics[width=0.18\textwidth]{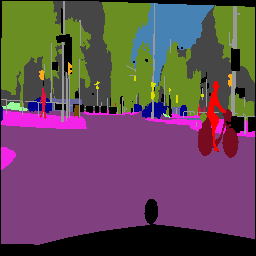}
         \includegraphics[width=0.18\textwidth]{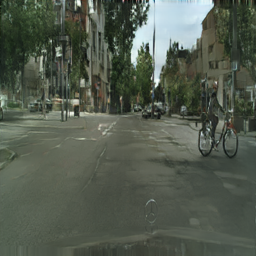}
         \includegraphics[width=0.18\textwidth]{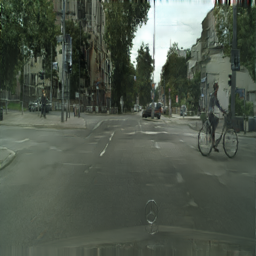} \\
        \smallskip
        \includegraphics[width=0.18\textwidth]{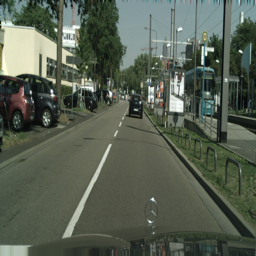}
         \includegraphics[width=0.18\textwidth]{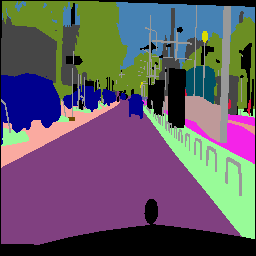}
         \includegraphics[width=0.18\textwidth]{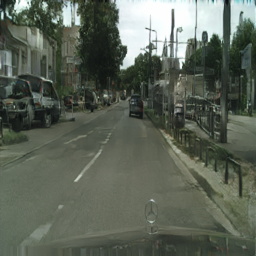}
         \includegraphics[width=0.18\textwidth]{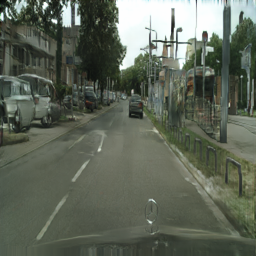} \\
         \smallskip
        \includegraphics[width=0.18\textwidth]{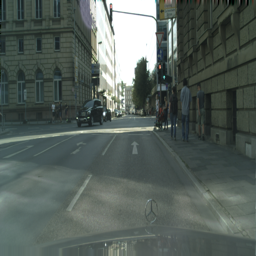}
         \includegraphics[width=0.18\textwidth]{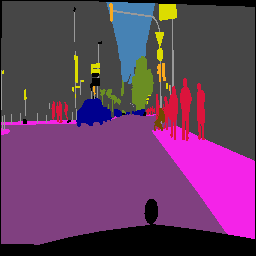}
         \includegraphics[width=0.18\textwidth]{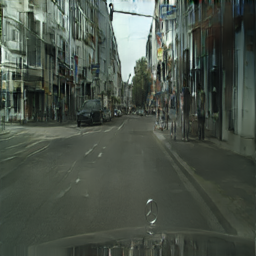}
         \includegraphics[width=0.18\textwidth]{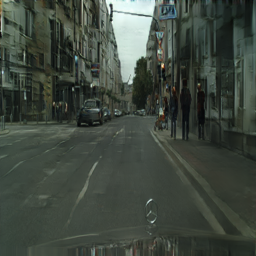} \\
    \end{center}
    \caption{Qualitative results of city scene image synthesis on Cityscapes dataset. Column (a) represents the ground truth. Its semantic map is shown in column (b). The results of the SPADE baseline using their pre-trained models are shown in column (c) followed by our proposed method in column (d).}
    \label{fig:city}
\end{figure*}

\begin{figure*}[!t]
    \begin{center}
        \makebox[0.18\textwidth][c]{(a) Ground truth}
		 \makebox[0.18\textwidth][c]{(b) Semantic input}
		 \makebox[0.18\textwidth][c]{(c) SPADE \cite{park2019gaugan}}
		 \makebox[0.18\textwidth][c]{(d) Ours} \\
		\smallskip
        \includegraphics[width=0.18\textwidth]{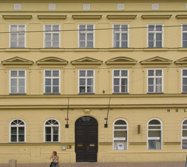}
         \includegraphics[width=0.18\textwidth]{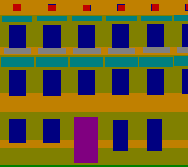}
         \includegraphics[width=0.18\textwidth]{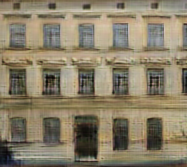}
         \includegraphics[width=0.18\textwidth]{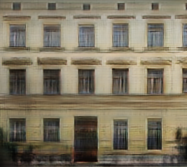} \\
        \smallskip
        \includegraphics[width=0.18\textwidth]{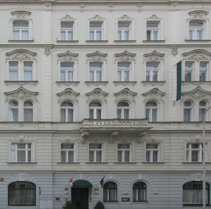}
         \includegraphics[width=0.18\textwidth]{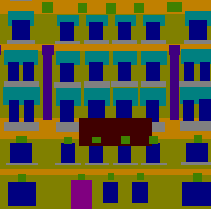}
         \includegraphics[width=0.18\textwidth]{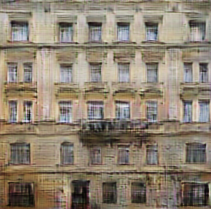}
         \includegraphics[width=0.18\textwidth]{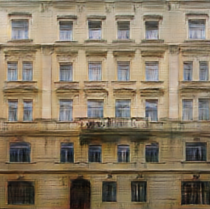} \\
        \smallskip
        \includegraphics[width=0.18\textwidth]{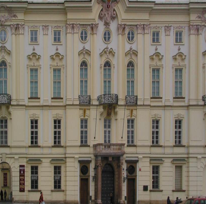}
         \includegraphics[width=0.18\textwidth]{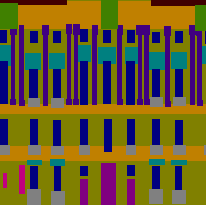}
         \includegraphics[width=0.18\textwidth]{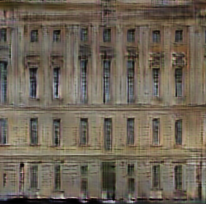}
         \includegraphics[width=0.18\textwidth]{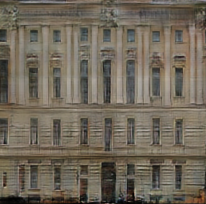}
    \end{center}
    \caption{Qualitative results of facade image synthesis on CMP Facades. Column (a) represents the ground truth. Its semantic map is shown in column (b). The results of SPADE baseline using their pre-trained models are shown in column (c) followed by our proposed method in column (d).}
    \label{fig:facades}
\end{figure*}

\paragraph{Qualitative results}
Qualitative results are shown in \Cref{fig:city} for Cistyscapes and in \Cref{fig:facades} for CMP Facades.
Having the semantic matching head and different feature maps, each focusing on a specific object, could generate more semantically consistent details, \textit{e.g.}, the windows and balconies are less blurry and with more details for the facades.
By giving equal weight to all classes, our generator is able to better synthesize small objects that have few pixels or that are less frequent, such as doors in Facades, and buses, bikes, trains, and baby strollers in Cityscapes.
The buildings are cleaner, and humans are more visible.

\paragraph{Quantitative results}

We report Fréchet Inception Distance (\textit{FID}) \cite{fid}, which compares the statistics of generated and real images.
The results are presented in the second column of \Cref{tab:city} and \Cref{tab:facades}.

Similar to previous works \cite{chen2017crn, park2019gaugan}, a pre-trained segmentation network is used to see how well the predicted semantic masks match the ground truth input.
In this study, Dilated Residual Network (DRN) \cite{yu2017drn} is used. The pre-trained models are obtained from their publicly available code.\footnote{\url{https://github.com/fyu/drn}} Note that we resize its input images to the resolution of $512 \times 256$.
The quantitative evaluations and the comparison with previous works are presented in \Cref{tab:city}. The results show an improvement in all per-pixel accuracy, per-class accuracy, and mean-IOU.

In order to have a visual fidelity comparison against previous works, the Amazon Mechanical Turk (AMT) was used. The workers are given a semantic input mask and the outputs of the methods and are asked to choose the image which is more matched to the mask and is more realistic. The experiment consists of $500$ questions, and each question is carried out by five different workers without any time limitation.
The results can be found in the last column of \Cref{tab:city}. It shows that users preferred our results more than the baseline.

\begin{table*}[!t]
    \centering
    \caption{Quantitative evaluation of scene synthesis on Cityscapes dataset. Image quality: FID (the lower the better). Segmentation performance: mIOU, pixel and class accuracies (the higher the better). User preference study: the numbers show how much people preferred that method.}
    \begin{tabular}{c|c|ccc|c}
        \toprule
        \multirow{2}{*}{Models} & \multicolumn{5}{c}{Cityscapes \cite{cordts2016cityscapes}} \\
        & FID $\downarrow$ & mIOU $\uparrow$ & pixel $\uparrow$ & class $\uparrow$ & user pref.\\
        \midrule
        Pix2pix \cite{isola2017pix2pix} & 79.1 & 28.5 & 67.2 & 29.0 & - \\
        Pix2pixHD \cite{wang2018pix2pixhd} & 67.8 & 35.8 & 83.9 & 43.5 & - \\
        Pix2pixHD + Ours & 55.8 & 44.4 & 89.2 & 52.7 & - \\
        ASAPNet \cite{shaham2021asapnets} & 69.2 & 29.6 & 77.2 & 35.1 & - \\
        ASAPNet + Ours & 57.3 & 42.1 & 88.6 & 50.1 & - \\
        \midrule
        SPADE \cite{park2019gaugan} & 56.8 & 47.0 & 90.1 & 54.7 & 40\% \\
        SPADE + Ours & \textbf{50.8} & \textbf{55.9} & \textbf{92.3} & \textbf{64.2} & \textbf{60\%}\\
        \bottomrule
    \end{tabular}
    \label{tab:city}
\end{table*}

\begin{table}[!t]
    \centering
    \caption{Quantitative evaluation of scene synthesis on CMP Facades dataset. Image quality: FID (the lower the better). User preference study: the numbers show how much people preferred that method.}
    \begin{tabular}{c|c|c}
        \toprule
        \multirow{2}{*}{Models} & \multicolumn{2}{c}{CMP Facades \cite{tylecek2013facades}} \\
        & FID $\downarrow$ & user preference \\
        \midrule
        SPADE \cite{park2019gaugan} & 121.4 & 29\% \\
        SPADE + Ours & \textbf{107.3} & 71\%\\
        \bottomrule
    \end{tabular}
    \label{tab:facades}
\end{table}

\paragraph{Evaluating on other baselines}
To show the generalization of our approach, we also applied the same procedure on pix2pixHD \cite{wang2018pix2pixhd} and recently presented ASAPNet \cite{huang2020semantic}. The results are in \Cref{tab:city}. To do that, we use the same generators as theirs and only modify their discriminators as described. We again observe how impactful our method is.
This modification of the discriminator of ASAPNet can improve their synthesized images without penalizing their speedup in inference since the discriminator is not used in inference time.

\subsection{Human synthesis from keypoints}
\label{sec:humans}

\paragraph{Datasets}
For the task of generating human images in a given pose (described as keypoints) with the same appearance as a source image, we validate our approach on the
DeepFashion dataset \cite{liu2016deepfashion}. This dataset includes $52,712$ clothing images with diverse person poses at the resolution of $256 \times 256$.
Similar to \cite{ma2017pg2}, $200,000$ pairs of the same person-clothes with two different poses are used.
We followed the same train/test split.
This dataset does not have pose information labels.
To obtain semantic annotations, a pre-trained pose detector \cite{cao2017openpose} with $K=18$ keypoints is used.
Results of Deformable GAN baseline \cite{siarohin2018deformable} are obtained from publicly available code.\footnote{\url{https://github.com/AliaksandrSiarohin/pose-gan}}

\paragraph{Implementation details}
The structure of $G$ is identical to Deformable GAN \cite{siarohin2018deformable}. We took both $G$ and $D$ from the baseline and modified its $D$.
Learning uses the same hyper-parameters as in \Cref{sec:scenes}, but for $100$ epochs.
Semantic masks $M_k(s_k)$ are obtained as gaussians centered on the keypoints with a variance of $\sigma = 6$.
Note that there is an extra encoder before the generator in this setting, which encodes the image appearance. This embedding vector is concatenated with the pose embedding and is fed to the decoder of the generator.

\begin{figure*}[!t]
    \begin{center}
        \makebox[0.11\textwidth][c]{\begin{tabular}{c} \scriptsize (a) Source\\\scriptsize image \end{tabular}}
		\makebox[0.11\textwidth][c]{\begin{tabular}{c}\scriptsize (b)  Target\\ \scriptsize image\end{tabular}}
		\makebox[0.11\textwidth][c]{\begin{tabular}{c}\scriptsize (c) Def.\\\scriptsize GAN \cite{siarohin2018deformable}\end{tabular}}
		\makebox[0.11\textwidth][c]{\scriptsize (d) Ours}
		\smallskip
		\makebox[0.11\textwidth][c]{\begin{tabular}{c}\scriptsize (a) Source \\\scriptsize image \end{tabular}}
		\makebox[0.11\textwidth][c]{\begin{tabular}{c}\scriptsize (b) Target\\\scriptsize image \end{tabular}}
		\makebox[0.11\textwidth][c]{\begin{tabular}{c}\scriptsize (c) Def.\\ \scriptsize GAN \cite{siarohin2018deformable}\end{tabular}}
		\makebox[0.11\textwidth][c]{\scriptsize (d) Ours} 
		\\
		\smallskip
        \includegraphics[trim={40 0 40 0}, clip,  width=0.11\textwidth]{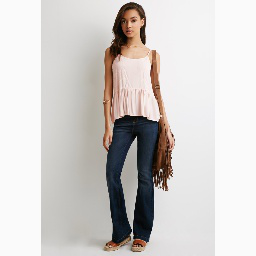}
        \includegraphics[trim={40 0 40 0}, clip, width=0.11\textwidth]{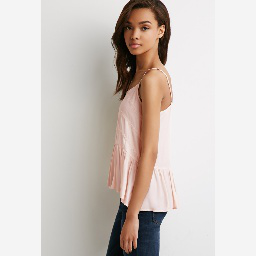}
        \includegraphics[trim={40 0 40 0}, clip, width=0.11\textwidth]{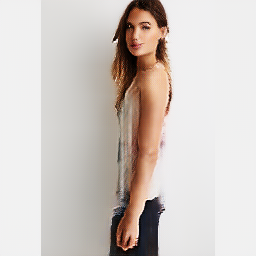}
        \includegraphics[trim={40 0 40 0}, clip, width=0.11\textwidth]{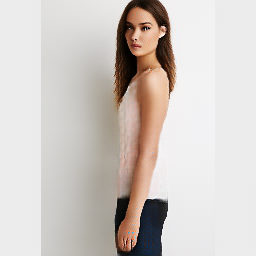}
        \smallskip
        \includegraphics[trim={40 0 40 0}, clip, width=0.11\textwidth]{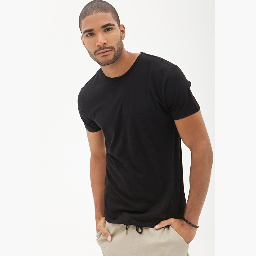}
        \includegraphics[trim={40 0 40 0}, clip, width=0.11\textwidth]{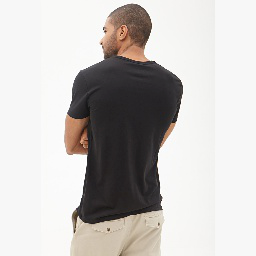}
        \includegraphics[trim={40 0 40 0}, clip, width=0.11\textwidth]{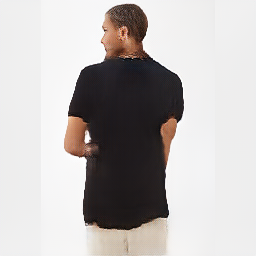}
        \includegraphics[trim={40 0 40 0}, clip, width=0.11\textwidth]{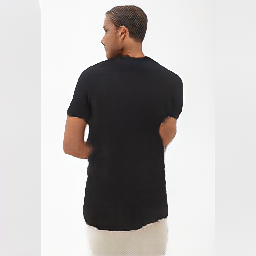} \\
        \smallskip
        \includegraphics[trim={40 0 40 0}, clip, width=0.11\textwidth]{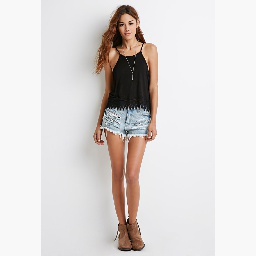}
        \includegraphics[trim={40 0 40 0}, clip, width=0.11\textwidth]{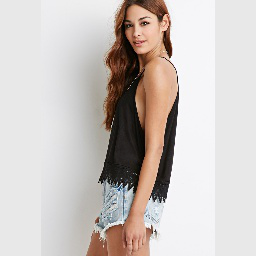}
        \includegraphics[trim={40 0 40 0}, clip, width=0.11\textwidth]{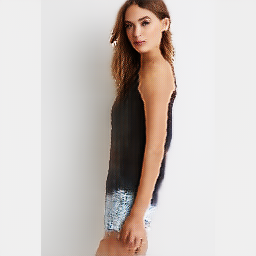}
        \includegraphics[trim={40 0 40 0}, clip, width=0.11\textwidth]{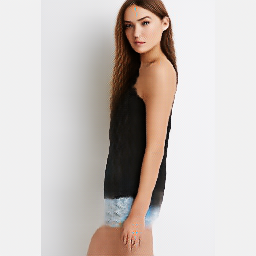}
        \smallskip
        \includegraphics[trim={40 0 40 0}, clip, width=0.11\textwidth]{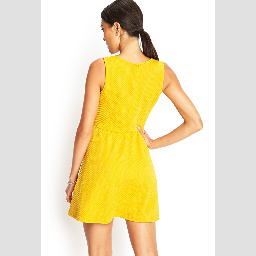}
        \includegraphics[trim={40 0 40 0}, clip, width=0.11\textwidth]{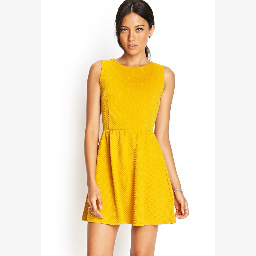}
        \includegraphics[trim={40 0 40 0}, clip, width=0.11\textwidth]{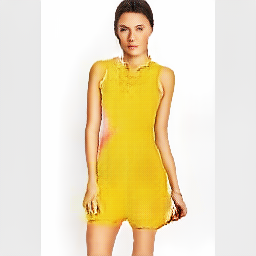}
        \includegraphics[trim={40 0 40 0}, clip, width=0.11\textwidth]{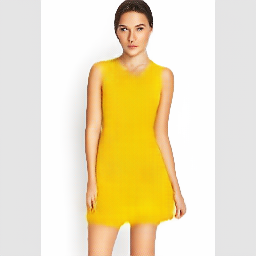}
    \end{center}
    \caption{Qualitative results of human image synthesis on DeepFashion. The source and target images are in columns (a) and (b) respectively.
    The baseline, Deformable GAN, is shown in column (c) followed by our proposed method in column (d).}
    \label{fig:deepfashion}
\end{figure*}

\paragraph{Qualitative results}
The results on the DeepFashion dataset are available in \Cref{fig:deepfashion}.
We observe that our discriminator adds more details, especially on faces and hands, without penalizing other parts.
These details are even more visible in high-resolution images.

\paragraph{Quantitative results}
Quantitative results are presented in \Cref{tab:scores_human}, where our proposed model achieves more realistic results in terms of FID than Deformable GAN.
This shows that overall, our generated images are closer to the real ones.
We have also performed a user preference study. We followed the same settings as the previous experiment.

\begin{table}[!t]
    \centering
     \caption{Quantitative evaluation of human image synthesis on DeepFashion.}
    \begin{tabular}{c|c|c}
        \toprule
        \multirow{2}{*}{Models} & \multicolumn{2}{c}{DeepFashion \cite{liu2016deepfashion}} \\
        & FID $\downarrow$ & user preference\\
        \midrule
        Deformable GAN \cite{siarohin2018deformable} & 125.25 & 23\% \\
        Deformable GAN + Ours & \textbf{106.27} & \textbf{77\%} \\
        \bottomrule
    \end{tabular}
    \label{tab:scores_human}
\end{table}

\subsection{Ablation study}

\begin{figure*}[!t]
    \begin{center}
        \makebox[0.24\textwidth][c]{\tiny{(a) Ground truth image}}
        \makebox[0.24\textwidth][c]{\tiny{(b) Baseline}}
		\makebox[0.24\textwidth][c]{\tiny{(c) Ours}}\\
        \includegraphics[width=0.24\textwidth]{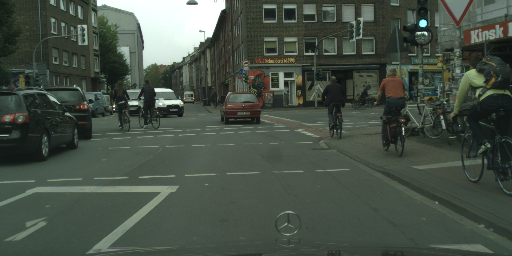}
        \includegraphics[width=0.24\textwidth]{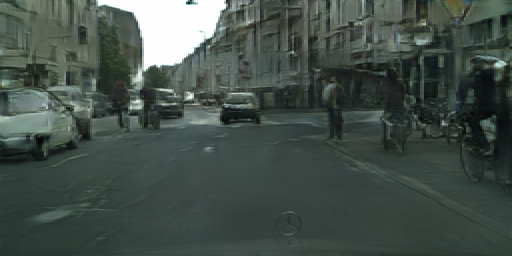}
        \includegraphics[width=0.24\textwidth]{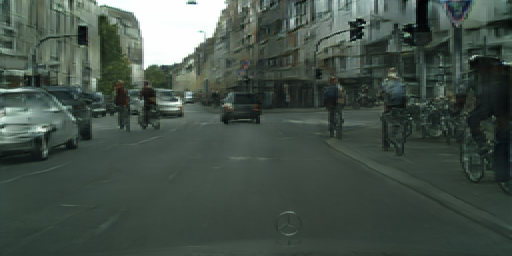}\\
        \includegraphics[width=0.24\textwidth]{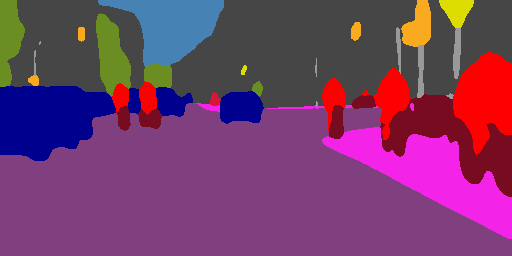}
        \includegraphics[width=0.24\textwidth]{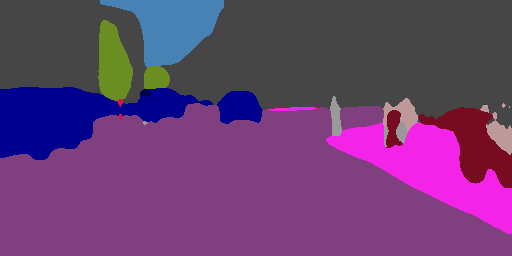}
        \includegraphics[width=0.24\textwidth]{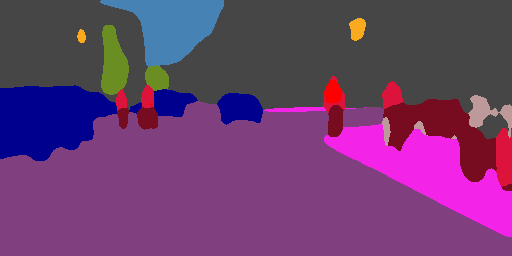}
    \end{center}
    \caption{Comparison of our model vs the baseline in terms of matching the condition semantic map. The inputs of the semantic segmentation model are shown in the first row, and its outputs are in the second row. We show the ground truth image (a), the synthesized image of the baseline (b) and ours (c). Ours respect the condition semantic map more.}
    \label{fig:visualization_semantics2}
\end{figure*}

\begin{figure*}[!t]
    \begin{center}
        \makebox[0.18\textwidth][c]{\tiny{(a) Input image}}
        \makebox[0.18\textwidth][c]{\tiny{(b) Ground truth semantic}}
		\makebox[0.18\textwidth][c]{\tiny{(c) $D_1$ output}}
		\makebox[0.18\textwidth][c]{\tiny{(d) $D_2$ output}} \\
		\smallskip
        \includegraphics[width=0.18\textwidth]{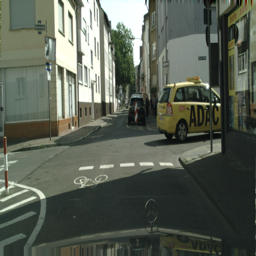}
        \includegraphics[width=0.18\textwidth]{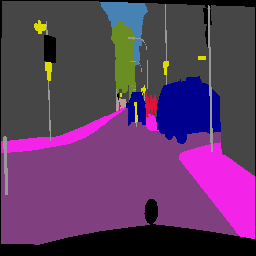}
        \includegraphics[width=0.18\textwidth]{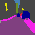}
        \includegraphics[width=0.18\textwidth]{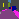}
    \end{center}
    \caption{Visualization of the semantic matching head outputs. The input of the discriminator is shown in (a) followed by the ground truth image (b). In the baseline, there are two discriminators for different scales and we output both of them (c, d) which (d) is in lower resolution.}
    \label{fig:visualization_semantics}
\end{figure*}

\label{sec:ablation}
In this section, an ablation study is provided to analyze the behavior of our model.
For the ablation study, we consider the scene synthesis task on Cityscapes and compare it to the above-defined baseline \cite{park2019gaugan}.

\paragraph{The effect of the perceptual loss} In this experiment, we want to observe the impact of the perceptual loss (extracted by a pretrained VGG network). As \Cref{tab:perceptual} shows, the performance of the baseline highly depends on the perceptual loss. However, the proposed method outperforms the baseline even without that loss. This proves that the proposed approach of using a multi-task semantic matching head could achieve better performance. Still, the perceptual loss is useful in improving the fidelity of images by adding some textures.

\paragraph{The effect of each head} The analysis of each head is demonstrated in \Cref{tab:ablation}. It shows that all three heads are effective.
The semantic matching head, with the help of the reconstruction task, has the best performance in segmentation metrics. Adding the coarse-to-fine adversarial head can effectively improve the quality of the images leading to a better FID.

\paragraph{Investigating the performance of the semantic matching head}
In \Cref{fig:visualization_semantics2}, the performance of the segmentation model on the synthesized images of our model is compared with the baseline. Humans, bikes, and traffic lights are clearly more detectable in ours.
We also visualize the semantic matching head output to see whether it correctly does the semantic extraction. To do that, the output of the discriminator for an image at evaluation time is depicted in \Cref{fig:visualization_semantics}. The baseline has two discriminators in different resolutions, and we show the outputs of both. 

\begin{table}[!t]
    \centering
    \caption{The comparison of the effect of perceptual loss in the baseline and ours. Here, our model has the semantic matching head, the reconstruction heads and a simple (not coarse-to-fine) adversarial head.}
    \begin{tabular}{c|c|ccc}
        \toprule
        \multirow{2}{*}{Models} & \multicolumn{4}{c}{Cityscapes \cite{cordts2016cityscapes}} \\
        & FID $\downarrow$ & mIOU $\uparrow$ & pixel $\uparrow$ & class $\uparrow$ \\
        \midrule
        SPADE \cite{park2019gaugan} w/o VGG & 63.0 & 42.1 & 88.6 & 49.7 \\
        SPADE \cite{park2019gaugan} w/ VGG & 56.8 & 47.0 & 90.1 & 54.7 \\
        Ours* w/o VGG & 56.2 & 54.8 & 92.2 & 62.4 \\
        Ours* w/ VGG & \textbf{52.6} & \textbf{56.3} & \textbf{92.3} & \textbf{63.9} \\
        \bottomrule
    \end{tabular}
    \label{tab:perceptual}
\end{table}

\begin{table}[!t]
    \centering
    \caption{Ablation study on the discriminator heads. c2f: coarse-to-fine adversarial head, rec: reconstruction head and sem: semantic matching head}
    \begin{tabular}{l|c|ccc}
        \toprule
        \multirow{2}{*}{Models} & \multicolumn{4}{c}{Cityscapes \cite{cordts2016cityscapes}} \\
        & FID $\downarrow$ & mIOU $\uparrow$ & pixel $\uparrow$ & class $\uparrow$ \\
        \midrule
        SPADE \cite{park2019gaugan} & 56.8 & 47.0 & 90.1 & 54.7 \\
        Ours (sem) & 53.4 & 55.9 & 92.2 & 63.7 \\
        Ours (sem + rec) & 52.6 & \textbf{56.3} & \textbf{92.3} & 63.9 \\
        Ours (c2f) & 52.7 & 45.2 & 89.7 & 52.5 \\
        Ours (c2f + sem) & 51.9 & 55.2  & 92.2 & 62.8 \\
        Ours (c2f + sem + rec) & \textbf{50.8} & 55.9 & \textbf{92.3} & \textbf{64.2} \\
        \bottomrule
    \end{tabular}
    \label{tab:ablation}
\end{table}

\paragraph{The effect of increasing the capacity of the network}
As previously mentioned, the added capacity to the architecture of $D$ is minimum. Only the last convolution layer of $D$ is modified, and instead of outputting a single feature map, it outputs multiple feature maps. The number of parameters for $D$ has increased by approximately $10\%$, and is fixed for $G$.
In one experiment, we increase the number of channels in the last layer of $D$ without inducing our approach. Thus, the number of parameters of $D$ is increased by $10\%$.
As \Cref{tab:capacity} shows, simply adding capacity without our shared representation could not improve the performance.
Moreover, note that the added capacity only affects the training time, and there is no overhead at inference time.

\begin{table}[!t]
    \centering
    \caption{The comparison of the effect of adding $10\%$ more capacity to the baseline vs ours. }
    \begin{tabular}{c|c|ccc}
        \toprule
        \multirow{2}{*}{Models} & \multicolumn{4}{c}{Cityscapes \cite{cordts2016cityscapes}} \\
        & FID $\downarrow$ & mIOU $\uparrow$ & pixel $\uparrow$ & class $\uparrow$\\
        \midrule
        SPADE \cite{park2019gaugan} & 56.8 & 47.0 & 90.1 & 54.7 \\
        SPADE + $10\%$ capacity & 57.9 & 44.5 & 89.8 & 52.1 \\
        Ours & \textbf{50.8} & \textbf{55.9} & \textbf{92.3} & \textbf{64.2} \\
        \bottomrule
    \end{tabular}
    \label{tab:capacity}
\end{table}

\section{Conclusions}
\label{sec:conc}
Image synthesis has a huge effect on modern transportation.
However, current models are not sufficiently photorealistic.
We presented a new semantically-aware discriminator to better guide the training of conditional  Generative  Adversarial  Networks. We showed that the task of the discriminator, classifying real/fake images, can be augmented with tasks closely related to understanding the content of the image at the pixel level. Our contributions are generic and can be applied to any generator network for image synthesis.
Future work will study extending our work to impose temporal consistency in video synthesis.

\section*{Acknowledgements.}
This project has received funding from the European union's Horizon 2020 research and innovation programme under the Marie Skłodowska-Curie grant agreement N$^{\circ}$ 754354, and Valeo.

\bibliographystyle{IEEEtran}  
\bibliography{references}

\end{document}